\documentclass[sigconf]{acmart}
\AtBeginDocument{%
  }

\copyrightyear{2025}
\acmYear{2025}
\setcopyright{cc}
\setcctype{by}
\acmConference[GeoAI '25]{The 8th ACM SIGSPATIAL International Workshop on AI for Geographic Knowledge Discovery}{November 3--6, 2025}{Minneapolis, MN, USA}
\acmBooktitle{The 8th ACM SIGSPATIAL International Workshop on AI for Geographic Knowledge Discovery (GeoAI '25), November 3--6, 2025, Minneapolis, MN, USA}
\acmDOI{10.1145/3764912.3770828}
\acmISBN{979-8-4007-2179-3/2025/11}

\usepackage{placeins}

\begin{document}

\title{Deep Learning to Identify the Spatio-Temporal Cascading Effects of Train Delays in a High-Density Network}

\author{Vu Duc Anh Nguyen}
\email{duc.anh.nguyen@outlook.de}

\affiliation{%
  \institution{University of Cologne}
  \city{Cologne}
  \state{NRW}
  \country{Germany}
}

\author{Ziyue Li}
\email{ziyue.li@tum.de}
\affiliation{%
  \institution{Technical University of Munich}
  \country{Germany}}

\renewcommand{\shortauthors}{Nguyen et al.}

\begin{abstract}
The operational efficiency of railway networks, a cornerstone of modern economies, is persistently undermined by the cascading effects of train delays. Accurately forecasting this delay propagation is a critical challenge for real-time traffic management. While recent research has leveraged Graph Neural Networks (GNNs) to model the network structure of railways, a significant gap remains in developing frameworks that provide multi-step autoregressive forecasts at a network-wide scale, while simultaneously offering the live, interpretable explanations needed for decision support.

This paper addresses this gap by developing and evaluating a novel XGeoAI framework for live, explainable, multi-step train delay forecasting. The core of this work is a two-stage, autoregressive Graph Attention Network (GAT) model, trained on a real-world dataset covering over 40\% of the Dutch railway network. The model represents the system as a spatio-temporal graph of operational events (arrivals and departures) and is enriched with granular features, including platform and station congestion. To test its viability for live deployment, the model is rigorously evaluated using a sequential, k-step-ahead forecasting protocol that simulates real-world conditions where prediction errors can compound.

The results demonstrate that while the proposed GATv2 model is challenged on pure error metrics (MAE) by a simpler Persistence baseline, it achieves consistently higher precision in classifying delay events---a crucial advantage for a reliable decision support tool. The explainability analysis, a key contribution, reveals a novel insight into the model's learned strategy: it dynamically focuses its attention on local station congestion (proxied by \texttt{Dwell} edges) to identify fragile network states where inter-train \texttt{Headway} conflicts are most likely to cause significant delay propagation. This work contributes a robust methodology for autoregressive forecasting on large-scale graphs, introduces the Edge Propagation Error metric for targeted evaluation, and provides a blueprint for a next-generation, explainable decision support system for railway operators.
\end{abstract}



\begin{teaserfigure}
  \includegraphics[width=\textwidth]{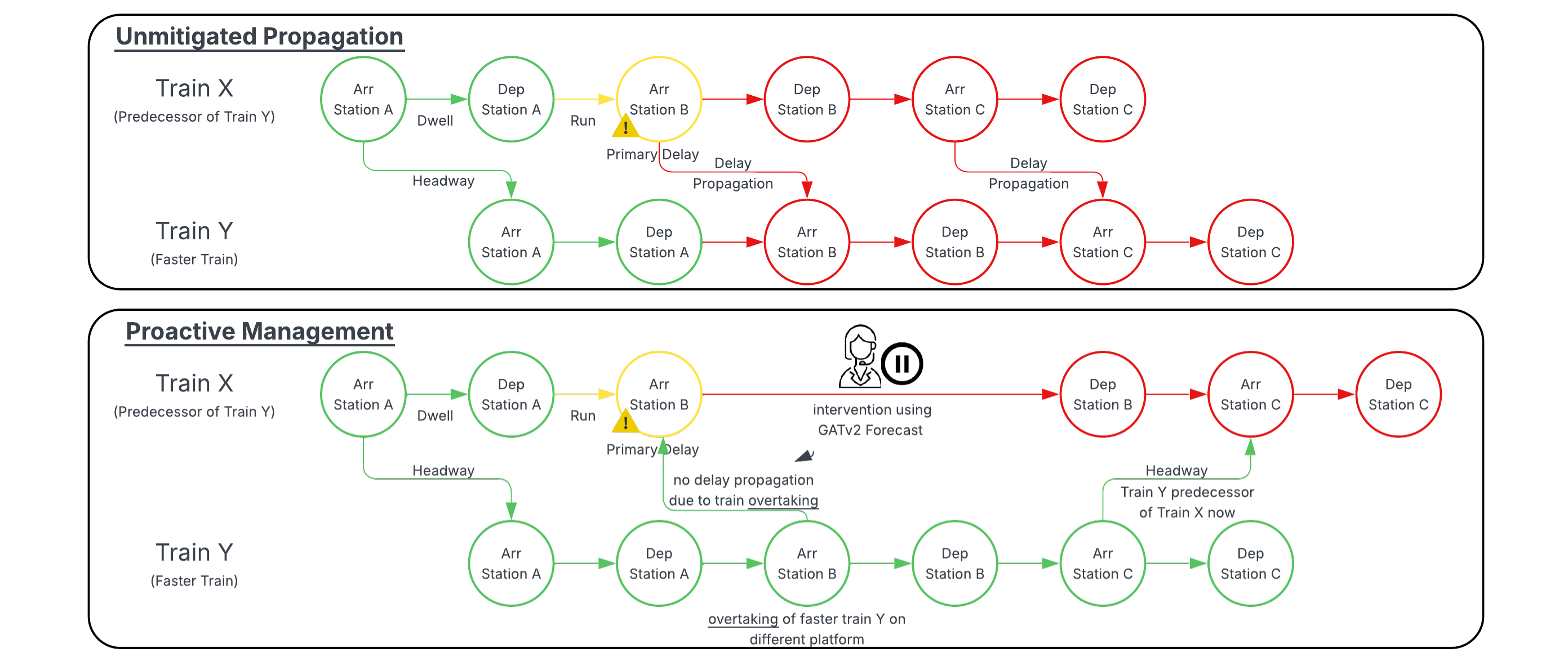}
  \caption{Conceptual diagram of the research problem and solution. Top: A primary delay on Train X cascades to Train Y. Bottom: A forecast from our GATv2 model enables a proactive intervention (overtaking) to prevent the secondary delay.}
  \Description{A two-panel flowchart. The top panel shows a delay, colored red, starting on Train X and spreading to Train Y. The bottom panel shows an intervention icon and Train Y overtaking Train X, with the delay propagation halted.}
  \label{fig:teaser}
\end{teaserfigure}

\maketitle

\section{Introduction}

Rail transport is a strategic asset for modern economies, enabling efficient freight logistics, supporting commuter mobility, and reducing transport-related emissions \cite{Association_of_American_Railroads_2025}. The European Union, for instance, identifies rail as a critical enabler of net-zero logistics due to its scalability and energy efficiency \cite{eu_rail_2025}. However, the operational efficiency of these geospatial networks is persistently undermined by service disruptions. Persistent delays disrupt supply chains, reduce commuter productivity, and erode public confidence, posing a significant economic and policy challenge \cite{Kecman2014, EC2020strategy}. The German network exemplifies this, with 2023 seeing over a third of long-distance services delayed, resulting in passenger compensation costs exceeding €200 million annually \cite{DB2023report, bnetza2024}.

The core operational challenge lies not in forecasting a singular delay, but in modeling the complex, systemic dynamics of its evolution through the network. Service disruptions are typically categorized into primary delays—initial, stochastic events like technical malfunctions—and secondary (or knock-on) delays, which are the cascading effects that spread throughout the network in a phenomenon known as delay propagation \cite{Wu2019Towards}. In densely scheduled networks, the tight coupling of services means a minor primary delay on one train can rapidly propagate to succeeding and interacting trains, amplifying the initial disruption \cite{Spanninger2022}. Therefore, the goal of advanced predictive models is to capture this systemic propagation, providing dispatchers with the tools to analyze the evolving state of the entire spatio-temporal system, estimate the risk of widespread disruption, and make informed decisions to mitigate these cascading effects. This necessitates a network-aware modeling approach that can move beyond isolated predictions to model the delay propagation process itself \cite{Huang2024Explainable}.

Historically, train delay prediction has evolved through several paradigms. Early approaches using classical statistical models or standard Artificial Neural Networks (ANNs) struggled to capture the complex non-linear and spatio-temporal interdependencies of a railway network, often focusing on single-train operations \cite{Spanninger2022, Li2022Prediction}. The advent of Graph Neural Networks (GNNs) represents a significant shift. The inherent graph structure of transportation systems makes GNNs a natural and powerful tool, as they are specifically constructed to learn rich representations that encode both network topology and spatial correlations \cite{Jiang2022}.

More recently, the research frontier has advanced from a focus on purely predictive accuracy to include Explainable Geospatial Artificial Intelligence (XGeoAI). Advanced models such as Graph Attention Networks (GATs) aim to uncover the underlying determinants of delay propagation, transitioning the core question from what a delay will be to why it is predicted to occur \cite{Huang2024Explainable}. This explainability is critical for building trust with human operators and enabling proactive decision-making, as identifying the key contributing factors can directly inform a dispatcher's intervention strategy.

Despite these significant advances, a critical research gap remains in developing a framework that can deliver robust, multi-step forecasts at a network-wide scale, while simultaneously providing the live, interpretable explanations needed for real-time decision support. Most state-of-the-art models are designed as single-step or one-shot predictors, treating forecasting as a retrospective analysis of a historical graph rather than a live, sequential simulation where prediction errors can compound over time \cite{Huang2024Explainable}. Furthermore, these models are often validated on smaller network corridors and rarely incorporate operative features such as platform and station congestion, which are critical drivers of secondary delays in dense operational environments.

To address this gap, this paper introduces a novel XGeoAI framework for live, explainable, and multi-step forecasting of train delay propagation. We represent the railway system as a dynamic spatio-temporal graph where nodes are discrete operational events (arrivals and departures) and edges represent their dependencies (running, dwelling, and headway). We propose a two-stage, deep autoregressive GAT model (specifically GATv2), rigorously tested on a large-scale, real-world dataset covering over 40\% of the highly dense Dutch passenger railway network and enriched with novel congestion features. Methodologically, this work extends beyond one-shot predictions by implementing a deep autoregressive simulation, enabling a rigorous test of a GAT's long-horizon stability and its robustness to compounding errors.

This framework is designed to investigate two central challenges that guide this work. The first concerns the forecasting horizon: to what extent can such a model reliably forecast knock-on delays at increasing k-step-ahead horizons under a live, sequentially consistent inference protocol? The second addresses explainability and edge-type contributions: how do different operational dependencies contribute to delay propagation, and can this be quantified by the model's learned attention weights? Our contributions are:

\begin{enumerate}
\item We design and evaluate a live, autoregressive forecasting simulation for a GAT-based model. Our results reveal a critical trade-off between raw predictive accuracy (MAE), where a simple persistence baseline excels, and the consistently higher precision our model achieves, a crucial advantage for a reliable decision support tool that must minimize false alarms.
\item We present a novel explainability finding that challenges simple interpretations of attention mechanisms. Our model learns a sophisticated, indirect strategy: it focuses its attention on local station congestion (identified via Dwell events) as a key signal to determine when delay propagation across inter-train Headway connections is most likely to occur.
\item We introduce the Edge Propagation Error (EPE), a metric designed for a more targeted evaluation of a model's ability to learn and predict the specific delay dynamics across different types of operational dependencies in the network.
\end{enumerate}

By providing a blueprint for a real-world decision support tool, our findings advance the application of XGeoAI for knowledge discovery and proactive management in dynamic, large-scale transportation networks (see Figure~\ref{fig:teaser}).

\section{Related Work}

The challenge of predicting train delays has seen a significant evolution in modeling paradigms. Early deep learning models, such as LSTMs, improved upon classical statistical methods but were fundamentally limited by their inability to naturally represent the irregular topology of railway systems, often focusing on single-train operations \cite{Wen2020Predictive, Huang2020CLFNet, bronstein2017geometric}. This led to the adoption of graph-based models that explicitly encode the operational dependencies between train events.

Initial graph-based approaches included "white-box" models like Bayesian Networks (BNs), which offered high interpretability by learning statistical dependencies from historical data \cite{Corman2018}. However, these models are often limited by simplifying assumptions, such as linear relationships between events, and their learned dependencies are static. This highlights a key limitation: while explainable, their explainability is not dynamic, failing to capture how the influence of one event on another changes with network conditions \cite{Spanninger2022}.

To overcome these limitations, the current research frontier is dominated by GNNs, which combine the structural advantages of graph representations with the power of deep learning to capture complex, non-linear spatio-temporal patterns. While some GNNs operate at an aggregated station level, they lose the fine-grained operational detail necessary to model propagation dynamics \cite{Zhang2022TSTGCN}. The most advanced methods, therefore, represent the system as a dynamic event graph, where each arrival and departure is a node, allowing for the explicit modeling of running, dwelling, and headway connections \cite{Li2024EAAG}.

Within this domain, the explainable GAT (X-GAT) \cite{Huang2024Explainable} represents the state-of-the-art. Structurally similar to our work, the X-GAT model uses an event-based graph and leverages the GAT's attention mechanism to provide dynamic, per-prediction explainability, successfully identifying headway edges as primary drivers of inter-train delay propagation. However, the X-GAT framework has two critical limitations that prevent its direct application as a live decision support tool. First, it treats forecasting as a retrospective, single-step graph regression task, predicting all delays in a historical subgraph simultaneously. This approach, while powerful for post-hoc analysis, does not address the challenge of real-time forecasting where prediction errors can compound over a multi-step horizon. Second, the model was validated on a limited scale—a single railway corridor over a two-month period—and did not incorporate granular, real-world features like platform and station congestion.

Our work is directly motivated by this combination of gaps. We extend the state-of-the-art by moving from a retrospective, one-shot analysis to a live, multi-step autoregressive forecasting approach. This more challenging but operationally realistic protocol allows us to rigorously test a GAT's long-horizon stability and its robustness to its own compounding errors. Furthermore, we validate our framework on a large-scale network with the inclusion of novel congestion features, addressing the critical need for models that are not only explainable but also scalable and robust enough for real-world deployment.

\section{Methodology}
\label{sec:methodology}

This section details the technical framework developed to address the identified research gaps. The primary objective is to construct and evaluate a model capable of serving as a real-time decision support tool for railway dispatchers by moving beyond static, one-shot predictions and developing a deep autoregressive forecasting approach \cite{janjos2023bridging}.

\subsection{Problem Formulation}
\label{sec:problem_formulation}

The primary task is to perform a multi-step forecast for a sequence of \(k\) future train events, predicting the delay magnitude for each. The model operates under a live forecasting scenario, where all past events are known. The railway system is formally defined as a directed spatio-temporal graph \(G = (V, E)\), where:
\begin{itemize}
    \item \textbf{Nodes (V):} Each node \(v \in V\) represents a single, discrete train event (an arrival or a departure) at a specific station, associated with a feature vector \(\mathbf{x}_v\).
    \item \textbf{Edges (E):} An edge \(e_{ij} \in E\) represents one of three key operational dependencies: Running Edges (connecting a departure to its subsequent arrival), Dwelling Edges (connecting an arrival to its departure at the same station), and Headway Edges (connecting events of two different trains that use the same infrastructure).
\end{itemize}
For any part of the graph representing events that have already occurred, the edge features contain the actual realized durations. For any future, part of the graph to be predicted, these edge features are populated with their scheduled durations from the timetable to respect sequential order.

Node features are categorized into derived timetable attributes (e.g., stop name, train type, peak hour flags, cancellation flags, stops remaining), dynamic features reflecting the network state (e.g., the lagged delay 2 stops prior), and station and platform congestion metrics.

\subsection{Data, Features, and Graph Construction}
\label{sec:data_and_graph}

\subsubsection{Data and Pre-processing}
The primary dataset consists of a comprehensive archive of all passenger train services in the Netherlands from 2019 onwards, provided by \textit{Rijden de Treinen} \cite{RijdenDeTreinen} and sourced from the official Dutch public transport data provider, NDOV \cite{NDOVloket}. The raw data, augmented with platform-level information, underwent extensive pre-processing to ensure its integrity. To handle services where a train number changes mid-trip, a unique trip identifier was created. The dataset was scoped to the core passenger network by removing non-standard services (e.g., rail replacement buses).

The primary focus was resolving temporal inconsistencies; given the small number of affected trips (typically <2\%), our strategy was to remove corrupt observations rather than impute them, preventing the injection of noise. This included eliminating records with illogical sequences (e.g., a departure before an arrival) and systematically handling partial and full cancellations to create a clean event sequence for each trip. Finally, to focus the analysis on the most delay-prone parts of the network while managing computational complexity, the dataset was filtered to trips passing through the Randstad metropolitan area, a dense region accounting for over 55\% of the total network delay \cite{Levkovich2016}. To properly represent the cyclical nature of time, arrival and departure times were transformed into cyclical features using sine and cosine transformations \cite{lazzeri2020machine}.

\subsubsection{Feature Engineering}
Following the cleaning process, a set of features was engineered to provide the model with a rich view of the network's state. Intra-train dynamics were captured by calculating the duration of running and dwelling activities, following the formal definitions used in state-of-the-art event-based models \cite{Huang2024Explainable}. For a train \(j\) at its \(i\)-th stop representing event \(v\), these are:
\[ \text{RunningDuration}(v_{dep}^{(j,i-1)}) = at(v_{arr}^{(j,i)}) - at(v_{dep}^{(j,i-1)}) \]
\[ \text{DwellingDuration}(v_{arr}^{(j,i)}) = at(v_{dep}^{(j,i)}) - at(v_{arr}^{(j,i)}) \]
where \(at(\cdot)\) is the actual, realized time of an event.

To model critical inter-train interactions, a sophisticated headway duration feature was engineered, defined as the time difference between the arrival events of two trains at the same station \cite{mcnaughton2011signalling}: \[ \text{HeadwayDuration}(v_{arr}^{(j,i)}) = at(v_{arr}^{(j,i)}) - at(v_{arr}^{(j',i)}) \]
To ensure operational relevance, a headway interaction was only considered if the two trains were scheduled within 30 minutes of each other and shared a common downstream path. This nuanced definition is designed to identify pairs of trains likely to enter into conflict over shared infrastructure, providing the model with a more accurate signal for learning how operational conflicts lead to delay propagation \cite{corman2012biobjective}.

To provide the model with essential temporal context, several time-based features were created. These include binary flags for the day of the week, weekend, and month. A holiday flag was also included, based on a definitive list of public holidays in the Netherlands for the operational year \cite{OfficeHolidays2022}. Furthermore, a flag for peak hours was engineered based on the official definitions provided by the Dutch Railways (NS), covering the morning rush hour (06:30-09:00) and the evening rush hour (16:00-18:30) \cite{NS2024Peak}.

To address a key gap in the literature, we engineered a set of novel features to explicitly quantify platform and station congestion. For an event \(v_i\) at station \(s\) and platform \(p\), we calculate:
\[ \text{PlatformCongestion}(v_i) = [N(p, t-60, t), \Delta t(p, t)] \]
where \(N(p, t-60, t)\) is the count of other train arrivals at platform \(p\) in the prior 60 minutes, and \(\Delta t(p, t)\) is the minutes since the last train arrived. Station congestion is defined as:
\[ \text{StationCongestion}(v_i) = M(s, t-5, t+5) \]
where \(M(s, t-5, t+5)\) is the count of all other train movements at station \(s\) within a \(\pm\)5-minute window. These features provide a measure of local traffic density, a critical driver of secondary delays \cite{lindfeldt2012congested}. Finally, all categorical features (e.g., train type, station name) were converted to learnable embeddings, and continuous features were log-transformed and standardized as needed \cite{kuhn2013applied, james2013introduction}.

\subsubsection{Graph Construction}
The final step is to assemble the engineered features into a graph structure. For each stop event in the cleaned dataset, two distinct nodes are instantiated: one for the arrival and one for the departure. A feature matrix \(\mathbf{X}\) and a target delay vector \(\mathbf{y}\) are then populated. These nodes are then connected by the three edge types (Dwell, Run, Headway) based on the pre-calculated operational dependencies. The resulting graph object includes node features, edge indices, edge attributes (e.g., actual running time), and a categorical edge type identifier. Crucially, scheduled timetable data is attached as metadata, as it allows the model to access the original timetable information to dynamically update the graph with its own predictions.

\subsection{Model Architecture and Training}
\label{sec:ModelAndTraining}

The development of the forecasting model followed a systematic process. An initial attempt to create a single, unified GATv2 model to solve both classification (will a delay occur?) and regression (how large will it be?) simultaneously proved to be fundamentally unstable during training. A systematic investigation revealed that the gradients for the classification task were consistently orders of magnitude larger than those for the regression task, indicating a task conflict within the shared layers.

Based on this finding, we pivoted to a two-stage model architecture, often referred to as a "Hurdle" model \cite{baetschmann2017dynamic}. This approach separates the competing tasks by employing two specialized models: a Classifier to first determine if a delay will occur, followed by a Regressor to predict the magnitude for only those events flagged by the classifier.

\subsubsection{Theoretical Foundation}
GNNs are a class of deep learning models designed to operate directly on graph-structured data. Their core mechanism is an iterative process called message passing, where each node aggregates information from its local neighborhood to update its own representation \cite{Wu2020Comprehensive}. We employ GATs, which enhance this process by using an attention mechanism to dynamically weigh the importance of different neighbors for each prediction \cite{Velickovic2018Graph}. Specifically, this work uses GATv2, a more expressive variant whose dynamic attention mechanism is crucial for explaining the context-dependent drivers of delay propagation, as the importance ranking of neighboring nodes is conditioned on the target node itself \cite{Brody2021HowAttentive}.

\begin{figure*}[htbp]
    \centering
    \includegraphics[width=\textwidth]{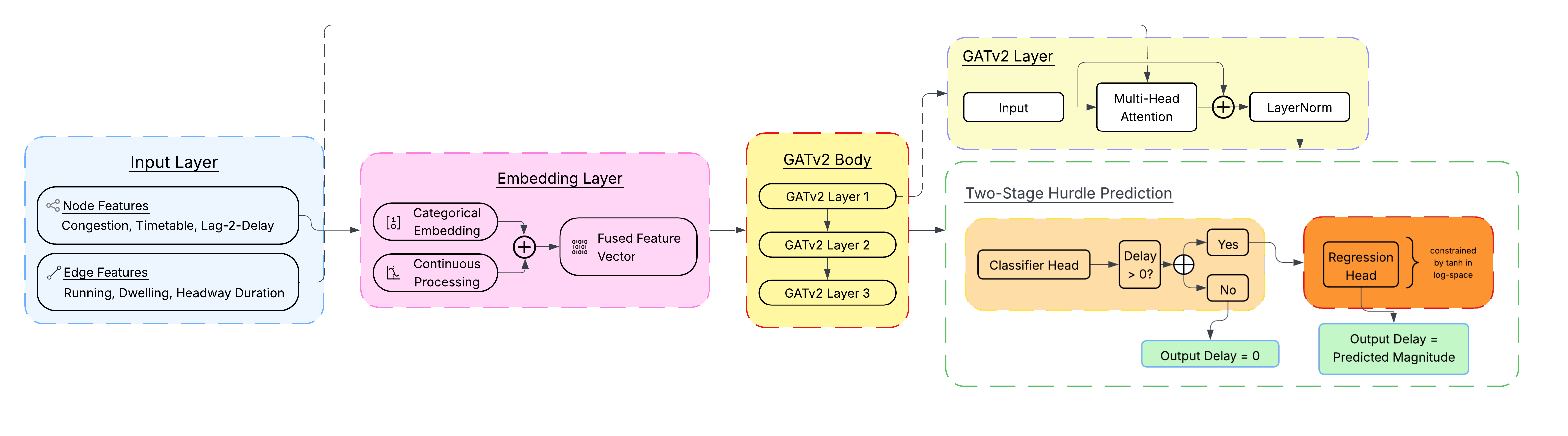}
    \caption{The complete two-stage GATv2 model architecture. Stage 1 \& 2 (Input \& Embedding Layer) processes raw node and edge features into fused vectors. Stage 3 (GATv2 Body) uses a stack of GATv2 layers with residual connections to learn graph representations. Stage 4 (Two-Stage Hurdle Prediction) splits the output into a Classifier and a Regressor and combines the output into a final delay forecast.}
    \Description{A detailed flowchart showing the four stages of the model. Raw inputs are processed and fused. They then pass through a three-layer GNN Body. A zoomed-in view of a GNN layer shows an input, a Multi-Head Attention block bypassed by a residual connection, a summation point, and a LayerNorm block. The GNN output splits into a Classifier head and a Regressor head, whose results are combined in a final prediction step.}
    \label{fig:model_architecture}
\end{figure*}

\subsubsection{Model Architecture: Two-Stage GATv2}
The final methodology is a two-stage model composed of two independent, architecturally similar GATs: a Classifier and a Regressor, as visualized in Figure~\ref{fig:model_architecture}. Both components share the same core GAT body to ensure they have an equal capacity to learn from the complex graph structure.

\paragraph{Core GAT Body} The core of each model is a deep Graph Attention Network. The main body is a stack of three GATv2Conv layers, each with 32 hidden channels and employing 32-head attention. To enable the effective training of a deep architecture and mitigate issues like vanishing gradients, residual connections are used to link the outputs of successive layers \cite{he2016deep}, and a LayerNorm is applied after each GATv2 layer to ensure stable training \cite{ba2016layer}.

\paragraph{Component 1: The Classifier}
The first component is a GAT model designed exclusively to solve the binary classification problem: will an event be delayed (delay > 0) or not? It uses the core GAT body described above, and its final node embeddings are passed to a single linear classification head which produces a raw logit for the prediction.

\paragraph{Component 2: The Regressor}
The second component is a GAT model designed for the specialized task of predicting the magnitude of a delay, conditioned on a delay occurring. It shares the same core GAT architecture as the classifier. Its final node embeddings are passed to a single linear regression head whose output is constrained by a \(5.0 \times \tanh(x)\) function to ensure stable, bounded predictions in log-space. This architectural choice prevents the exploding predictions observed in early experiments by bounding the output to a range of approximately 147 minutes that covers the vast majority of observed delays.

\subsubsection{Autoregressive Training and Forecasting}
The two-stage model is trained using a deep autoregressive approach designed to simulate a live forecasting scenario where the model must be robust to its own compounding errors. A key challenge in training such models is exposure bias: a model trained exclusively using ground-truth data (a technique known as teacher forcing) is never exposed to its own errors and can become biased during a live forecast where it must use its own, potentially imperfect, predictions as input \cite{NIPS2016_16026d60}. To mitigate this, we employ scheduled sampling, a learning strategy that gradually shifts the model from using ground-truth data to using its own predictions as input during training \cite{NIPS2015_e995f98d}. This teaches the model to be robust to its own compounding errors.

The Classifier and Regressor are trained in two completely independent, back-to-back simulations for each service day. Both follow a k-step rollout structure.
The Classifier's goal is to learn the boundary between delayed and on-time events. At each step \(s\) of its rollout, a sequentially consistent subgraph is extracted for the target events. The model performs inference, and the Binary Cross-Entropy (BCE) loss is calculated against the true binary labels (delay > 0). The state update for the next step, \(s+1\), is designed to simulate a realistic future: a model-driven state is created where the true delay value is used if the classifier predicts a delay, and a zero delay is used otherwise. A scheduled sampling strategy then chooses between this model-driven state and the pure ground-truth state to update the graph's edge attributes and historical features for the next step in the simulation.

Immediately after, the Regressor begins its own simulation on the same daily graph. Its specialized goal is to learn the magnitude of delays, conditioned on one having occurred. At each step \(s\), it outputs a constrained, log-space delay prediction. The Mean Squared Error (MSE) loss is then calculated against the true log-delay. Critically, a loss mask is applied, ensuring that MSE is only calculated for events that were actually delayed. This architecturally ensures the regressor is never penalized for its predictions on on-time events and only learns from actual delays. The state update for the regressor is more direct: scheduled sampling selects either the model's predicted log-delay or the true log-delay to update the graph's state for the next step.

\subsubsection{Forecasting Approach}
\label{sec:ForecastingApproach}
The forecasting process is a robust and sequentially consistent simulation of a live deployment. It strictly avoids data leakage from the future by using an autoregressive rollout where the model's own predictions construct the input features for subsequent steps. The pre-trained models are loaded and set to evaluation mode. The test dataset is constructed by randomly sampling 30 distinct service days.

The approach iterates chronologically through each service day. For a batch of anchor events, the k-step simulation begins with teacher forcing completely disabled. At each step \(s\), the live feature matrix is updated; for instance, the lagged delay feature is populated using predictions made two steps prior. A minimal, sequentially consistent subgraph is extracted to ensure the model only sees information from the realized past.

\begin{table*}[t]
\caption{Overall Model Performance Comparison}
\label{tab:overall_performance}
\centering
\begin{tabular}{lrrrrrr}
\toprule
\textbf{Model} & \textbf{MAE} & \textbf{RMSE} & \textbf{Accuracy} & \textbf{Precision} & \textbf{Recall} & \textbf{F1-Score} \\
\midrule
GATv2 & 1.0927 & 2.7432 & \textbf{0.6896} & \textbf{0.6229} & 0.4572 & 0.5274 \\
GCN & 1.0828 & 2.5946 & 0.6850 & 0.5899 & \textbf{0.5524} & \textbf{0.5705} \\
Zero Delay & 1.2597 & 3.4787 & 0.6213 & 0.0000 & 0.0000 & 0.0000 \\
Persistence & \textbf{0.9734} & \textbf{2.4282} & 0.6858 & 0.6059 & 0.4875 & 0.5403 \\
\bottomrule
\end{tabular}
\end{table*}

The core of the logic is the two-stage hurdle prediction. First, the Classifier is run on the subgraph. If its prediction is `False` (no delay), the final predicted delay is set to 0.0 minutes. If its prediction is `True`, the same subgraph is fed into the Regressor, which outputs a precise, log-space prediction for the delay's magnitude. This final prediction (either 0.0 or the real-scale value from the regressor) is then used to update the simulation state for the next step, \(s+1\). This includes dynamically re-calculating headway edge durations based on the new predictions. This process repeats for k steps.

\subsection{Evaluation Framework}
\label{sec:evaluation_framework}

The model's accuracy is assessed against three carefully selected baseline models, each chosen to test a specific aspect of the proposed framework. A Zero-Delay baseline (always predicts 0) establishes a naive benchmark, while a Persistence baseline (predicts the last known delay 2 stops prior for all future steps) is used to verify that the model learns complex propagation dynamics rather than just copying a recent state.

The most significant baseline is a two-stage, one-shot Graph Convolutional Network (GCN) model, designed to isolate the contributions of our key architectural choices. This GCN baseline tests two central hypotheses: by replacing GATv2Conv layers with simpler GCNConv layers and employing a one-shot forecasting strategy instead of a sequential rollout, it directly quantifies the performance gain attributable to the attention mechanism and our autoregressive approach. The GCN baseline maintains the two-stage structure but uses standard GCNConv layers and \(k\) independent output heads to generate all predictions in a single forward pass.

Performance is quantified using standard regression (MAE, RMSE) and classification (Precision, Recall, F1-Score) metrics \cite{zheng2015evaluating}. We prioritize Precision, as a low rate of false positives is critical for a trusted decision support tool. To assess robustness, performance is also evaluated across several operational subgroups, including by delay magnitude, operational context (peak vs. off-peak), train type, and at geographic hotspots (busiest stations).

To provide a novel, quantitative measure of the model's learned dynamics, we introduce the EPE. For an edge connecting an upstream event \(v_i\) to a downstream event \(v_j\), EPE is formally defined as:
\[ \epsilon_{\text{prop}}(v_i, v_j) = |(\text{delay}(v_j) - \text{delay}(v_i)) - (\widehat{\text{delay}}(v_j) - \widehat{\text{delay}}(v_i))| \]
where \(\text{delay}(\cdot)\) is the true delay and \(\widehat{\text{delay}}(\cdot)\) is the predicted delay. The MAE of this error is computed separately for Run, Dwell, and Headway edges. This allows us to directly evaluate the model's fidelity in capturing the specific delay dynamics of each operational link.

Finally, to provide qualitative support for our explainability claims, we analyze the GAT's learned attention weights to identify the characteristics of the most influential edges \cite{ying2019gnnexplainer}. This is complemented by a global Permutation Feature Importance analysis to assess the influence of each input feature on the model's predictions \cite{Breiman2001}.

\section{Results and Discussion}
\label{sec:results_and_discussion}

This section presents and interprets the empirical results of the experiments, structuring the findings to systematically address the research questions. We synthesize the quantitative results with their qualitative implications to provide a holistic analysis.

\subsection{Overall Performance: MAE-Precision Trade-off}
The primary evaluation of our proposed two-stage GATv2 model reveals a nuanced but critical trade-off in performance when compared against the baseline models, as summarized in Table~\ref{tab:overall_performance}.

As shown in Table~\ref{tab:overall_performance}, the results reveal a challenging performance landscape. The much simpler Persistence baseline, which predicts future delays based on the last known delay two stops prior, achieves the best scores on the key regression metrics (MAE and RMSE). While our GATv2 model does not achieve the lowest overall error, its value is demonstrated in its consistently higher precision compared to all baselines. For a real-world decision support tool, this is a critical advantage. Minimizing false alarms (high precision) is often more important for building operator trust and ensuring the adoption of a new system than marginal gains in average error. This finding is further emphasized by the one-shot GCN model, which achieved a higher overall F1-score by adopting a high-recall strategy at the cost of significantly lower precision, making it more prone to false alarms.

\subsection{Forecasting Horizon and Model Reliability}
\begin{figure}[h!]
    \centering
    \includegraphics[width=\columnwidth]{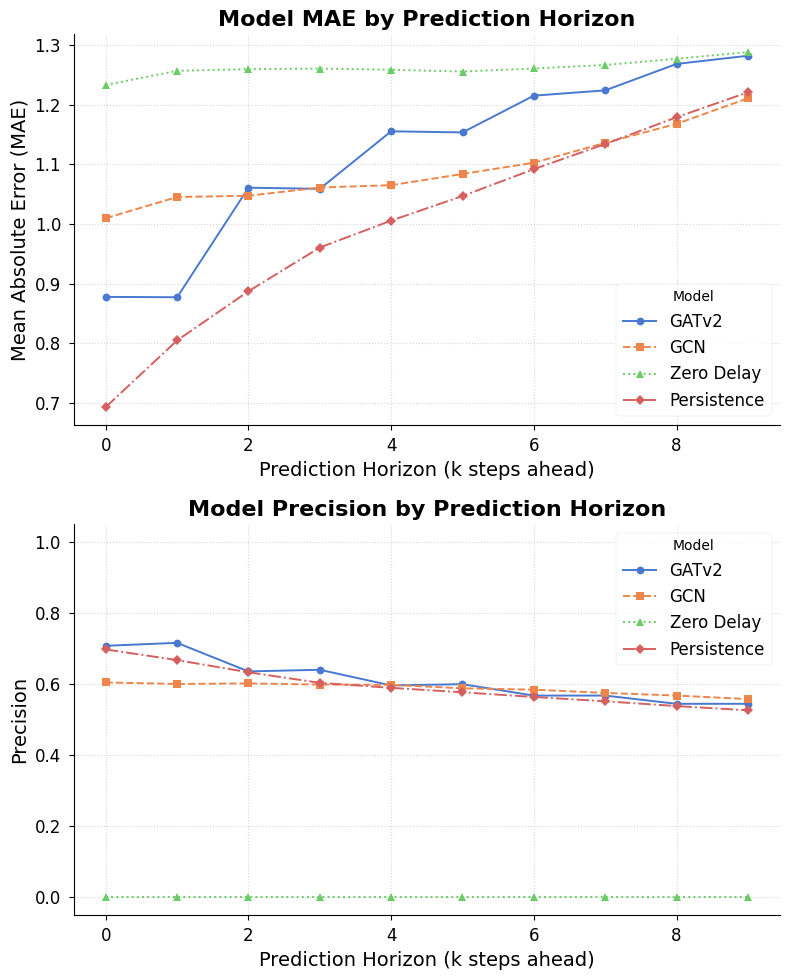}
    \caption{Model performance degradation over the k-step forecast horizon. Top: MAE. Bottom: Precision.}
    \Description{Two stacked line charts showing performance versus forecast step k (0–9). The top panel plots mean absolute error (MAE); MAE increases with k for all models. Persistence attains the lowest MAE across most horizons, while GATv2 rises from about 0.88 at k=0 to about 1.15 at k=4. The bottom panel plots precision; all models decline with horizon, but GATv2 remains the highest or near-highest, dropping from roughly 0.72 at k=0 to about 0.55 at k=4. Zero Delay stays near zero precision. Axes: x = forecast step k, y = MAE (top) or precision (bottom).}
    \label{fig:horizon_performance}
\end{figure}

This trade-off becomes even clearer when analyzing performance over the k-step forecast horizon, as shown in Figure~\ref{fig:horizon_performance}. This analysis directly addresses our first research question concerning the model's ability to provide reliable forecasts at increasing propagation horizons.

\begin{table}[t]
\centering
\caption{Permutation Feature Importance. Importance is the drop in model performance when a feature is shuffled.}
\label{tab:app_feature_importance}
\small
\begin{tabular}{lr}
\toprule
\textbf{Shuffled Feature} & \textbf{MAE Importance} \\
\midrule
lag2\_delay & 1.4449 \\
arrival\_tod\_cos & 0.0196 \\
minutes\_since\_last\_train\_clipped & 0.0140 \\
arrival\_tod\_sin & 0.0139 \\
is\_origin\_stop & 0.0108 \\
train\_count\_last\_60min & 0.0106 \\
number\_of\_stops\_left & 0.0073 \\
prev\_stop\_cancelled & 0.0013 \\
inbound\_trains\_near\_arrival & 0.011 \\
num\_prev\_cancelled & 0.0006 \\
is\_weekend & 0.0002 \\
platform\_known & -0.0000 \\
is\_peak & -0.0001 \\
is\_holiday & -0.0004 \\
platform\_scheduled & -0.0012 \\
day\_of\_week & -0.0029 \\
train\_type & -0.0062 \\
month & -0.0078 \\
is\_terminal\_stop & -0.0113 \\
stop\_name & -0.0327 \\
\bottomrule
\end{tabular}
\end{table}
As expected, the predictive performance degrades for all data-driven models as the forecast horizon \(k\) increases. As seen in the top of Figure~\ref{fig:horizon_performance}, the MAE for the GATv2 model rises from approximately 0.88 at \(k=0\) to 1.15 at \(k=4\). Notably, the Persistence baseline maintains a lower MAE than our GATv2 model for the majority of the forecast horizon. The feature importance analysis (see Table~\ref{tab:app_feature_importance}) provides a crucial insight into this: our model's predictions are overwhelmingly dependent on the lagged delay feature, suggesting a heavy reliance on a persistence-like mechanism rather than a full exploitation of the graph structure. This also explains the rapid degradation in performance after the first few steps. However, the bottom of Figure~\ref{fig:horizon_performance} shows that the GATv2 model consistently achieves the highest or near-highest precision, starting at 0.72 and degrading to approximately 0.55. This reinforces its value as a more reliable, albeit not always the most accurate, forecasting tool for a decision support context.

\subsection{Explainability: An Indirect Strategy for Propagation}
The most significant findings of this work come from the analysis of the model's internal dynamics, which address our second research question by revealing a sophisticated and non-obvious strategy for predicting delay propagation.

\subsubsection{Edge Propagation Error}
To quantify the model's ability to learn the specific dynamics of each operational link, the EPE was analyzed. Table~\ref{tab:classification_epe} evaluates how well each model can predict whether a delay will change across an edge.

\begin{table}[h!]
\caption{EPE Classification Performance by Edge Type}
\label{tab:classification_epe}
\centering
\footnotesize
\setlength{\tabcolsep}{3pt}
\begin{tabular}{l|ccc|ccc}
\hline
& \multicolumn{3}{c|}{\textbf{GATv2}} & \multicolumn{3}{c}{\textbf{GCN}} \\
\textbf{Edge Type} & \textbf{F1} & \textbf{Prec.} & \textbf{Recall} & \textbf{F1} & \textbf{Prec.} & \textbf{Recall} \\
\hline
Run & 0.5057 & 0.4750 & 0.4478 & \textbf{0.5705} & 0.4258 & \textbf{0.7051} \\
Dwell & 0.4417 & 0.1838 & 0.1160 & \textbf{0.5538} & 0.1181 & \textbf{0.5538} \\
Headway & \textbf{0.6001} & 0.6118 & 0.6241 & 0.5979 & 0.6165 & \textbf{0.6392} \\
\hline
\end{tabular}
\end{table}

The results reveal important trade-offs. Our GATv2 model is most effective at identifying propagation events across inter-train Headway edges, achieving the highest F1-Score (0.60) in that category. This confirms they are the primary mechanism for delay transfer between trains. In contrast, the GCN model achieves a higher F1-Score on Run and Dwell edges, driven by very high recall at the expense of significantly lower precision.

\subsubsection{Learned Attention Mechanisms}
\begin{figure}[htbp]
    \centering
    \includegraphics[width=\columnwidth]{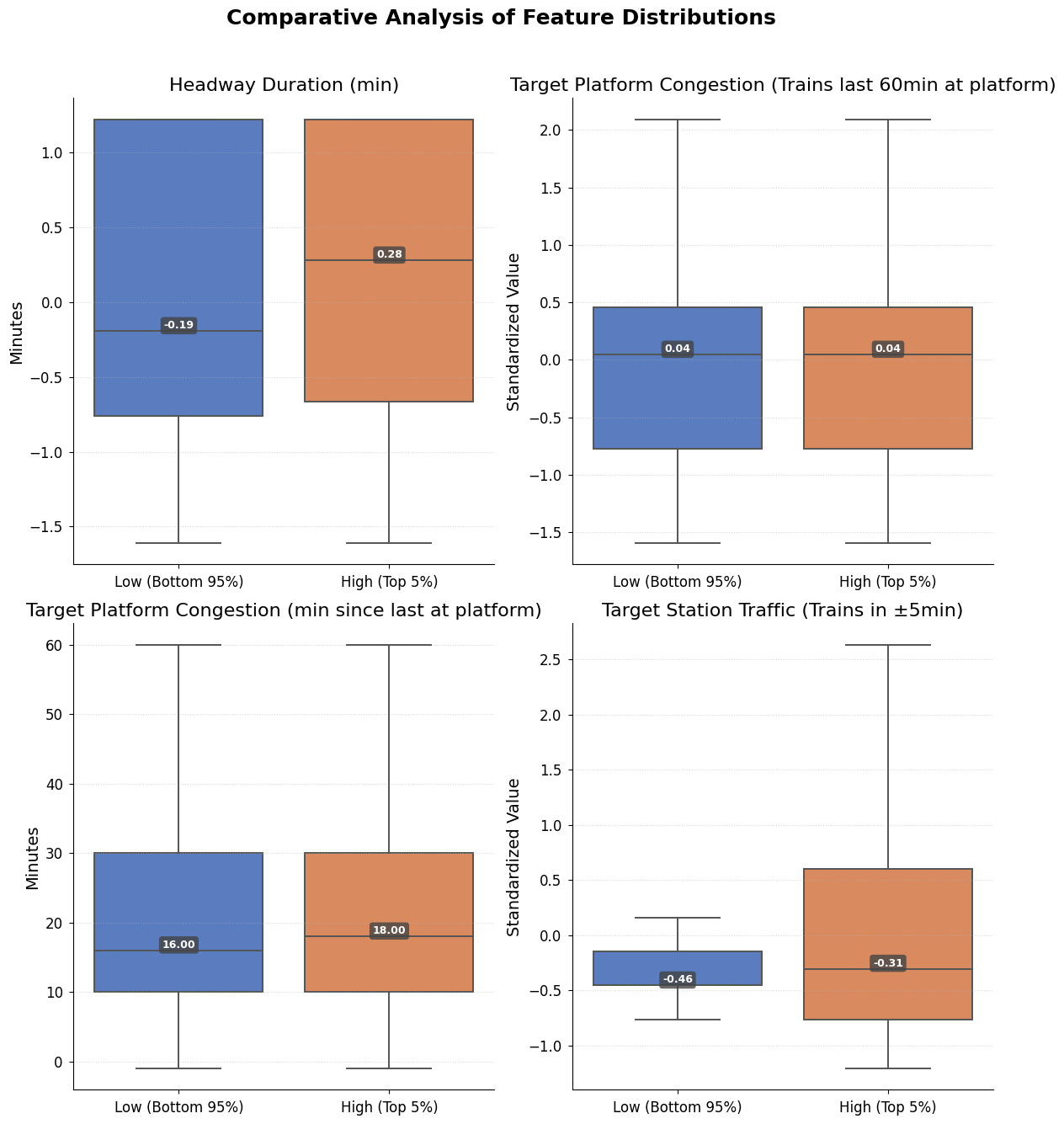}
    \caption{Comparative analysis of feature distributions for low vs. high-attention edges.}
    \label{fig:app_attention_features}
    \Description{Four box plots comparing feature distributions for low vs. high-attention edges. The plots show that high-attention edges are associated with higher station and platform congestion.}
\end{figure}

\begin{figure}[h!]
    \centering
    \includegraphics[width=\columnwidth]{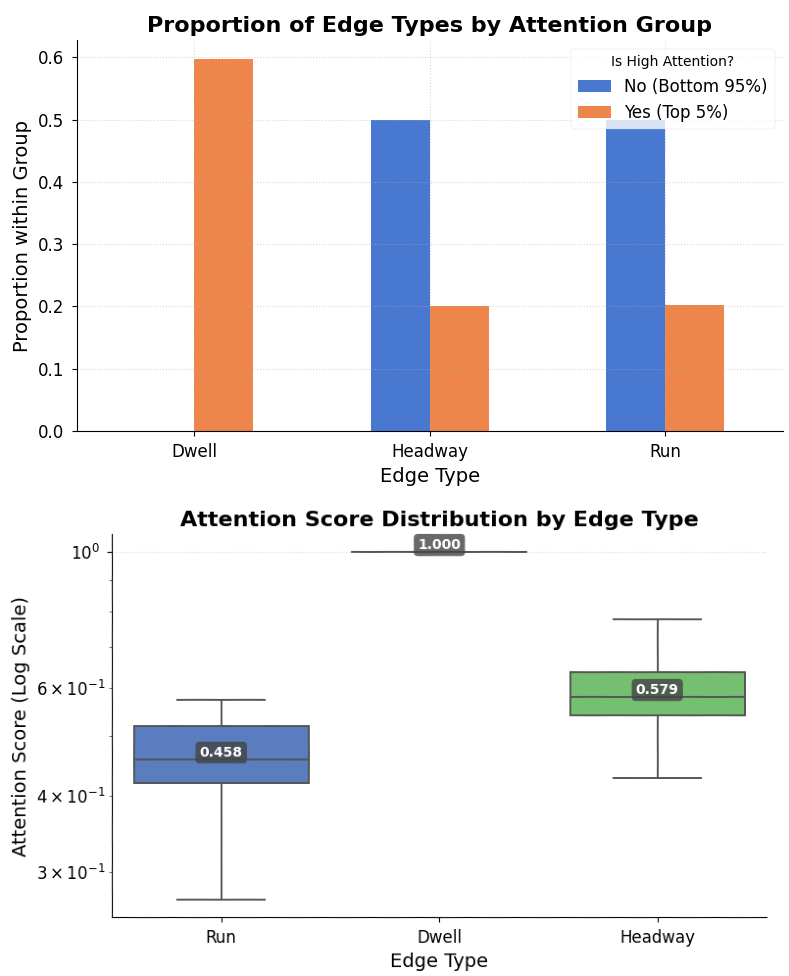}
    \caption{Analysis of GATv2 attention scores. Top: Proportion of Edge Types in High-Attention Group. Bottom: Distribution of Attention Scores by Edge Type.}
    \label{fig:attention_analysis}
    \Description{A two-part figure analyzing GATv2 attention, stacked vertically. The top bar chart shows that Dwell edges make up the largest proportion of high-attention edges. The bottom box plot shows that Dwell edges have the highest median attention score.}
\end{figure}

Interestingly, the analysis of the GAT's learned attention scores, shown in Figure~\ref{fig:attention_analysis}, reveals that the model does not focus most intensely on Headway edges, as might be expected given their importance in delay transfer. Instead, as shown in the top of Figure~\ref{fig:attention_analysis}, Dwell edges are overwhelmingly dominant in the high-attention group, accounting for nearly 60\% of the top 5\% most influential edges. This is confirmed in the bottom of the figure, where the median attention score for Dwell edges is 1.0. Further analysis in Figure~\ref{fig:app_attention_features} indicates that this high attention on Dwell edges is strongly correlated with periods of higher station and platform congestion. This finding directly addresses our second research question and challenges a simple interpretation of attention as a direct proxy for propagation, suggesting a more complex reasoning process.

\subsubsection{Interpretation of the Learned Strategy}
This compelling divergence between where the model focuses its attention (Dwell edges) and where propagation is most accurately predicted (Headway edges) suggests the model has learned a crucial operational insight. It uses local congestion (measured via Dwell events) as a critical signal to identify high-risk, fragile states of the network where conflicts between trains (Headway events) are most likely to cascade into significant delays. This demonstrates a sophisticated, indirect reasoning strategy that goes beyond simple, direct correlations. While our novel congestion features were found to be less important than the historical delay signal in the global feature importance analysis, their clear role in guiding the model's attention mechanism confirms their value in contributing to the model's explainability and addressing a key gap in the literature.

\subsection{Performance Across Subgroups}
To assess the model's robustness, its performance was evaluated across several operational subgroups. A key finding is that prediction error for all models increases significantly with the magnitude of the actual delay, with the largest errors occurring on severely delayed trains. In contrast, model performance remains largely consistent across different operational contexts (peak vs. off-peak hours, weekdays vs. weekends). An analysis by train type reveals that the models perform best on frequent commuter services like the Sprinter and struggle most with long-distance and international services. Geographically, the highest prediction errors were observed at major international hubs like Schiphol Airport and Amsterdam Centraal. (Full results are available in Appendix~\ref{app:subgroups}).

\subsection{Limitations and Future Work}
While this work successfully demonstrates a novel framework for explainable delay forecasting, it is important to acknowledge its limitations, each of which provides a clear avenue for future research. The development was constrained by computational resources, limiting hyperparameter tuning. The decision to use a logarithmic transformation and a `tanh` activation to stabilize the regressor, while necessary, introduced a systematic underestimation of large delays. While the GATv2 architecture was chosen for its explainability, it is crucial to interpret the resulting attention scores with caution; the mechanism highlights correlation, not necessarily causation \cite{baibing}, and its utility in explainability is tied to the model's predictive accuracy. Finally, this static graph representation does not account for dynamic operational changes like real-time re-routing, and the model's heavy reliance on the lagged delay feature suggests it may not be fully exploiting the rich topological information encoded in the graph.

This work lays a strong foundation for several exciting research directions. A crucial next step is to enrich the feature set by integrating external data sources, such as real-time weather, disruption information, or passenger load data to better model dwell times. Methodologically, a valuable extension would be to implement a GAT-based one-shot baseline or other ablation studies (e.g. removing headway edges to quantify impact of local congestion) to better isolate the specific impact of the autoregressive approach. Furthermore, exploring more advanced architectures like Relational Graph Attention Networks could allow the model to explicitly learn the distinct semantics of run, dwell, and headway edges. Finally, future work could explore more sophisticated techniques for handling the zero-inflated nature of delay data, potentially leading to more accurate and stable regression performance.

In summary, the empirical results provide nuanced answers to our initial research questions. The investigation into the forecasting horizon revealed a critical trade-off between the raw predictive accuracy of simpler models and the superior precision our GATv2 model offers, a key requirement for a trusted decision support tool. Concurrently, our explainability analysis uncovered a novel, indirect reasoning strategy learned by the model, where it leverages local congestion to predict inter-train propagation. By passing messages across headway edges that connect consecutive train events at the same physical location, the model learns how spatial resource conflict influences inter-train delay propagation. These findings confirm the viability of our autoregressive framework and contribute new insights into how GATs can achieve geographic knowledge discovery in complex, real-world networks, directly addressing the gaps in live forecasting and dynamic explainability.

\section{Conclusion}
\label{sec:conclusion}

This paper confronted a critical gap in the literature by developing a framework for live, multi-step, and explainable forecasting of train delay propagation in a large-scale railway network. The primary contribution was the design and rigorous evaluation of an autoregressive GATv2 model that leverages features like station congestion to provide not just predictions, but actionable insights for traffic management.

Our research successfully demonstrated the viability of this approach, providing nuanced answers to our guiding research questions. In response to the challenge of the forecasting horizon, the study confirmed that while our autoregressive GATv2 model is challenged on error magnitude by a simple Persistence baseline, it offers a crucial advantage in its superior precision, making it a more reliable tool for a real-world decision support system. Answering the question of explainability, this work uncovered a novel, indirect strategy learned by the GAT: it focuses its attention on local station congestion (via Dwell edges) as a key precondition for identifying when delay propagation is most likely to occur across inter-train Headway connections.

Methodologically, this work advances the field by implementing a live autoregressive simulation and introducing the EPE metric for a more targeted evaluation. Empirically, it provides critical insights into the challenges of scaling GATs and the complex trade-offs between model architecture, explainability, and predictive performance. Ultimately, this paper provides a blueprint for a next-generation, explainable decision support tool and demonstrates that the synergy of deep learning and graph-based modeling provides a powerful paradigm for not only predicting the future of a railway network but truly understanding its complex, interconnected dynamics.

\bibliographystyle{ACM-Reference-Format}
\bibliography{main}

\appendix



\section{Subgroup Performance Analysis}
\label{app:subgroups}

This section provides a detailed breakdown of model performance across several distinct operational subgroups to assess robustness and identify potential biases.

\paragraph{Performance by Delay Magnitude}

\begin{figure}[htbp]
    \centering
    \includegraphics[width=\columnwidth]{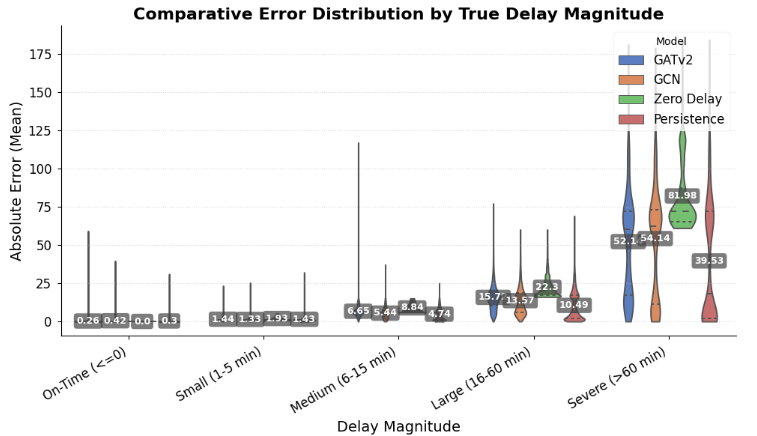}
    \caption{Comparative error distribution by true delay magnitude.}
    \label{fig:app_error_by_magnitude}
    \Description{A violin plot showing that the absolute error for all models increases as the true delay magnitude increases, with the largest errors and widest distributions for severe delays over 60 minutes.}
\end{figure}

Figure~\ref{fig:app_error_by_magnitude} shows that the prediction error for all models increases significantly with the magnitude of the actual delay. The largest errors and widest error distributions occur on severely delayed trains (>60 minutes), indicating that predicting the exact magnitude of large, infrequent delays remains a major challenge.

\paragraph{Performance by Operational Context}

\begin{figure}[htbp]
    \centering
    \includegraphics[width=0.8\columnwidth]{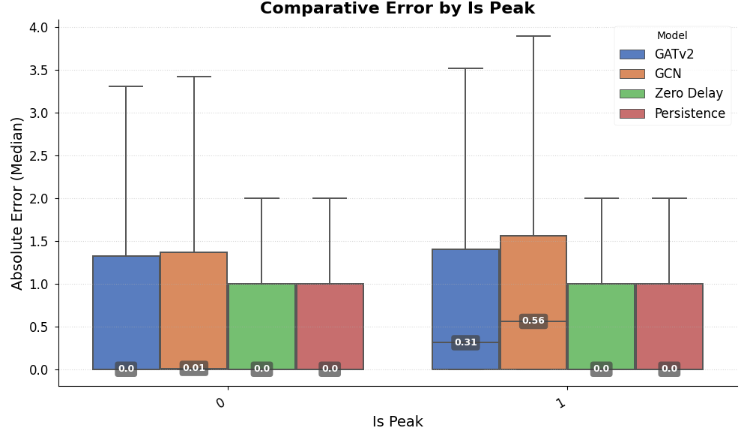}
    \caption{Model Error by Peak vs. Off-Peak Hours.}
    \label{fig:app_error_peak}
    \Description{A bar chart showing similar median absolute error for all models during both peak and off-peak hours, indicating consistent performance.}
\end{figure}

\begin{figure}[htbp]
    \centering
    \includegraphics[width=0.8\columnwidth]{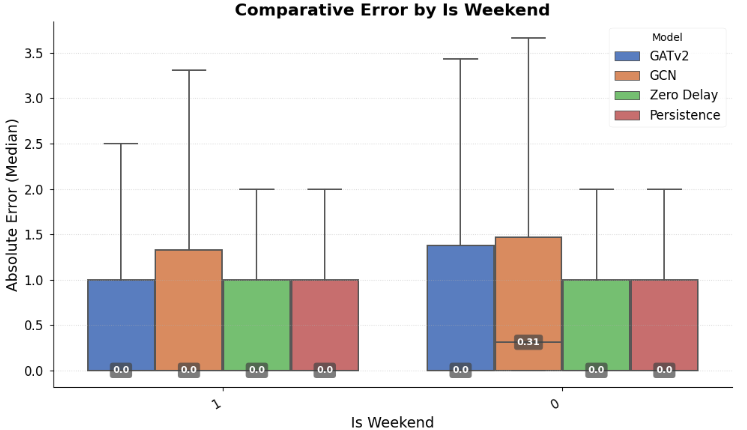}
    \caption{Model Error by Weekend vs. Weekday.}
    \label{fig:app_error_weekend}
    \Description{A bar chart showing similar median absolute error for all models during both weekends and weekdays, indicating consistent performance.}
\end{figure}

\begin{figure}[htbp]
    \centering
    \includegraphics[width=0.8\columnwidth]{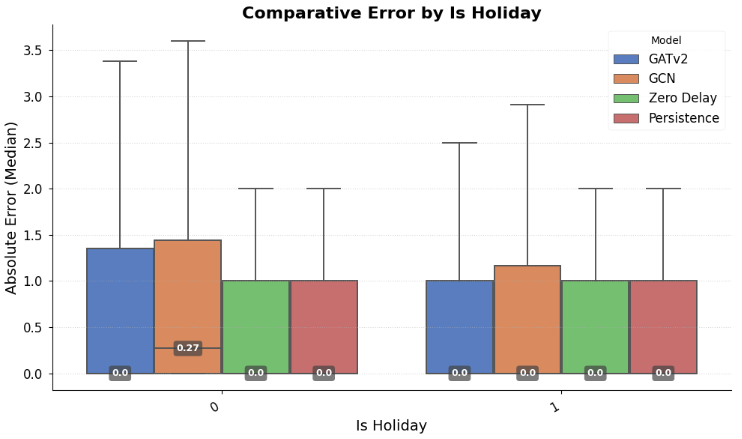}
    \caption{Model Error by Holiday vs. Non-Holiday.}
    \label{fig:app_error_holiday}
    \Description{A bar chart showing similar median absolute error for all models during both holidays and non-holidays, indicating consistent performance.}
\end{figure}

Figures~\ref{fig:app_error_peak}, \ref{fig:app_error_weekend}, and \ref{fig:app_error_holiday} show that model performance remains largely consistent across different operational contexts. There are only minor variations in error between peak and off-peak hours, weekdays and weekends, or holidays and regular days, suggesting the models are robust to these high-level temporal patterns.

\FloatBarrier

\paragraph{Performance by Train Type and Location}

The following tables break down performance by train type and geographic location. Table~\ref{tab:app_error_by_train_type} shows that all models perform best on frequent commuter services (like Sprinter) and struggle most with long-distance and international services that are less frequent and more prone to large, anomalous delays. Table~\ref{tab:app_error_by_station} shows that the highest prediction errors were observed at major international hubs like Schiphol Airport and Amsterdam Centraal, which handle complex operations.

\begin{table}[htbp]
\centering
\caption{Comparative Performance by Train Type (MAE in minutes).}
\label{tab:app_error_by_train_type}
\small
\begin{tabular}{lrrr}
\toprule
\textbf{Train Type} & \textbf{GATv2} & \textbf{GCN} & \textbf{Persistence} \\
\midrule
European Sleeper & 36.13 & 40.63 & 30.14 \\
Eurostar & 9.40 & 10.42 & 9.35 \\
Extra trein & 3.43 & 3.23 & 2.83 \\
ICE International & 8.54 & 9.59 & 7.24 \\
Int. Trein & 20.96 & 23.28 & 18.09 \\
Intercity & 1.32 & 1.38 & 1.23 \\
Intercity direct & 3.85 & 4.15 & 3.78 \\
Nachttrein & 4.06 & 3.05 & 3.04 \\
Nightjet & 23.78 & 27.82 & 21.50 \\
Speciale Trein & 0.52 & 0.50 & 0.53 \\
Sprinter & 0.90 & 0.84 & 0.76 \\
\bottomrule
\end{tabular}
\end{table}

\FloatBarrier

\begin{table}[htbp]
\centering
\caption{Comparative Performance at Top 10 Busiest Stations (MAE in minutes).}
\label{tab:app_error_by_station}
\small
\begin{tabular}{lrrr}
\toprule
\textbf{Station Name} & \textbf{GATv2} & \textbf{GCN} & \textbf{Persistence} \\
\midrule
Utrecht Centraal & 0.81 & 0.81 & 0.89 \\
Amsterdam Centraal & 1.56 & 1.71 & 1.40 \\
Amsterdam Sloterdijk & 1.20 & 1.17 & 1.03 \\
Rotterdam Centraal & 1.48 & 1.38 & 1.34 \\
Leiden Centraal & 1.10 & 1.10 & 1.11 \\
Den Haag Centraal & 1.12 & 1.02 & 0.99 \\
Schiphol Airport & 1.96 & 2.00 & 1.92 \\
Gouda & 0.85 & 0.89 & 0.84 \\
Delft & 1.47 & 1.34 & 1.19 \\
Amersfoort Centraal & 1.04 & 1.16 & 0.97 \\
\bottomrule
\end{tabular}
\end{table}

\end{document}